\pdfoutput=1

\documentclass[11pt]{article}

\usepackage[]{acl}

\usepackage{times}
\usepackage{latexsym}
\usepackage{CJKutf8}
\usepackage{csquotes}
\usepackage{graphicx}
\usepackage{float} 

\usepackage[T1]{fontenc}

\usepackage[utf8]{inputenc}

\usepackage{microtype}

\usepackage{inconsolata}
\usepackage{hyperref}

\title{Towards Better Inclusivity: A Diverse Tweet Corpus of English Varieties}

\author{
    Nhi Pham$^*$ \\
    New York University\\
    \texttt{nhi.pham@nyu.edu} \\\And
    Lachlan Pham$^*$ \\
    New York University \\
    \texttt{lp2233@nyu.edu} \\\And
    Adam Meyers \\
    New York University\\
    \texttt{meyers@cs.nyu.edu} \\
}

\begin{document}

\maketitle

\def\thefootnote{*}\footnotetext{These authors contributed equally to this work}\def\thefootnote{\arabic{footnote}}

\begin{abstract}
The prevalence of social media presents a growing opportunity to collect and analyse examples of English varieties. Whilst usage of these varieties was – and, in many cases, still is – used only in spoken contexts or hard-to-access private messages, social media sites like Twitter provide a platform for users to communicate informally in a scrapeable format. Notably, Indian English (Hinglish), Singaporean English (Singlish), and African-American English (AAE) can be commonly found online. These varieties pose a challenge to existing natural language processing (NLP) tools as they often differ orthographically and syntactically from standard English for which the majority of these tools are built. NLP models trained on standard English texts produced biased outcomes for users of underrepresented varieties \cite{DBLP:journals/corr/BlodgettO17}. Some research has aimed to overcome the inherent biases caused by unrepresentative data through techniques like data augmentation or adjusting training models. 

We aim to address the issue of bias at its root - the data itself. We curate a dataset of tweets from countries with high proportions of underserved English variety speakers, and propose an annotation framework of six categorical classifications along a pseudo-spectrum that measures the degree of standard English and that thereby indirectly aims to surface the manifestations of English varieties in these tweets. Following best annotation practices, our growing corpus features 170,800 tweets taken from 7 countries, labeled by annotators who are from those countries and can communicate in regionally-dominant varieties of English. Our corpus highlights the accuracy discrepancies in pre-trained language identifiers between western English and non-western (i.e., less standard) English varieties. We hope to contribute to the growing literature identifying and reducing the implicit demographic discrepancies in NLP. Code and dataset available on \href{https://github.com/phamleyennhi/diverse_english_corpus}{Github}.
\end{abstract}

\section{Introduction}

In many respects, there is a lack of linguistic and geographic diversity in NLP research. Whether that be in authorship identity, conference location, or target language \cite{faisal2021dataset}, it can be seen that wealthy, English-speaking nations are overrepresented. Even within these overrepresented nationalities, there exist subgroups, usually correlated with socioeconomic factors, who are not fairly represented \cite{blodgett-etal-2020-language}. A natural consequence is that the international proliferation and rapid advancement of language technologies serve the interests of the privileged.  

One major influence on the fairness of modern NLP systems is data availability. Our research targets the geographic diversity of NLP datasets and specifically seeks to introduce a new corpus that captures a range of English varieties. English’s status as a global lingua franca - as a consequence of historical colonialism and the current social influence of the western world - has meant that it has been adopted as an official language or, at least, it has become widely spoken in many countries across the world \cite{jenkins09}. In each culture, English takes on a different flavor to match the identities and needs of its speakers. In some cases, English is blended with other regional languages, forming pidgins and creoles.\footnote{This assumes a model of pigeons and creoles in which there is a shared dominant lexifier (e.g., English) and a diverse linguistic community that develops a shared language using modified words from the lexifier and other syntactic and phonological features from other languages, e.g., Caribbean languages \cite{handbook-pidgin-creole-studies}} In others, English takes over as a dominant language and phonetic indicators like accents or remnantal discourse markers become the main differentiators for that variety of English. Sometimes, English becomes a complementary language, children learning it alongside one or more other distinct languages, regularly code-switching according to contextual circumstances or to take advantage of nuances in connotation of different languages \cite{kortmann2008varieties}; each manifestation evinces a unique narrative about a speaker’s cultural and socio-economic background.  With such a range of Englishes spoken, it seems a missed opportunity that widely-used English data sets significantly overrepresent samples from America and Great Britain relative to their respective proportional constitution of global English speakers \cite{faisal2021dataset}.

As a direct consequence, language technologies are predominantly trained with western varieties of standard English and make algorithmic assumptions based on standard English grammar. Since the usage of English varieties are coupled tightly with geography and demography, there are numerous social and ethical implications of unrepresentative data. \newcite{https://doi.org/10.48550/arxiv.2105.05041} demonstrates there are accuracy biases in state-of-the-art speech recognition tools favoring accents and dialects most prevalent in training corpora. This reduces the effectiveness of, e.g., auto-captioning tools reducing the accessibility of global spoken media. \newcite{DBLP:journals/corr/BlodgettO17} shows that language identification performs worse on AAE than white-aligned English. This is notable because language identification is often an early filtration step in online text processing pipelines. For example, sentiment analysis research may discard anything labeled non-English including the higher proportion of false negatives for tweets of English varieties.

This research aims to address this need to diversify data by producing a corpus of tweets posted from English-speaking countries across Asia and Africa, as well as English tweets from the United States and the United Kingdom. Our work leverages the availability of geo-tagged and informal written texts provided by Twitter to create a dataset permeated by conversational dialectal features - features which were historically limited to the spoken medium. Furthermore, the recruited annotators come from the international student body of New York University which consists of multilingual individuals. Thus, these annotators are capable of extracting insights drawn from their experience in communities where a select English variety is commonplace. We assess the linguistic diversity of our corpus by examining the distribution of tweets across 6 categories and frequently-used words or phrases in each location. We also apply common language identifiers such as langid.py, spaCy, and Google Translate API to highlight their performance disparities compared to tweets of non-western English varieties. 

In section \ref{related_work}, we examine related NLP work on underserved English varieties, outline some other attempts at addressing unrepresentative corpora and describe some suggested best-practice guidelines that motivated our annotation process. Section \ref{data} elaborates on the choice of Twitter as a data source, explains the data collection approach and provides a brief linguistic variation analysis on our data. Section \ref{annotation} describes the annotation process including: a summarised set of annotation guidelines and its justification (section \ref{guidelines}), inter-annotator agreement benchmarks (section \ref{benchmarks}) and the recruitment and training of annotators (section \ref{training}). Section \ref{stats} evaluates the resulting corpus. Finally, section \ref{future_work} discusses potential future applications of the corpus and sets forth the imminent plans for the continuation of this research. In Section \ref{langid}, we investigate biases in several off-the-shelf language identification tools towards non-western English varieties. Section \ref{future_work} discusses potential applications and the continuation of this research. We also acknowledge several limitations of our corpus.

\section{Related Work} \label{related_work}
\subsection{Corpus Development}
The need to address the underrepresentation of English varieties in corpora has been recognised and approached in a number of ways ranging from manual collection to automatic generation. 
\newcite{dacon-etal-2022-evaluating} develops a rule-based method to translate standard American English (SAE) sentences to African-American English (AAE). This tool - CodeSwitch - relies on a list of 20 deterministic substitutions that aim to preserve “textual accuracy i.e., the original structure, intent, semantic equivalence and quality of a text.” These are derived from the translations and annotations of 3 AAE speaking workers. An independent set of AAE speakers generally believed that the machine generated output of the tool resembled human translations. The researchers note that a deterministic substitution method lacks contextual awareness and, consequently, some of the machine generated text were indeed identified as most likely machine generated. Such a tool simultaneously overgeneralises and undergeneralises the use of AAE. Not only would the rules be reflective of the patterns of the 3 AAE-speaking workers, it assumes that every instance of a ‘translatable’ sequence would be converted into an AAE form. Similarly, the substitutions lack the inclusion of more complex varietal features such as idiomatic phrases or regional-specific lingo. Nevertheless, CodeSwitch is a highly scalable and rapid way to augment current SAE datasets with an English variety, leveraging most of the existing labels since substitutions preserve the overall syntactic qualities of the text and thereby reducing the labor-demanding need to create and label new datasets. 

The International Corpus of English (ICE) project \cite{greenbaum1996international} relies on the efforts of global linguistic research teams to produce English corpora for regions and countries where English is an official first or additional language. They curate and annotate various spoken and written texts like press editorials, news broadcasts and classroom lessons. This process is significantly labour-intensive and requires specialised local researchers. These corpora are much more suited towards qualitative sociolinguistic investigations due to the limited sample sizes. Furthermore, each corpus is composed predominantly of formal texts so most comparative observations would be made on the standard national varieties of English as opposed to the more dialectal and informal varieties that our investigation is targeting. 

\newcite{cook2017building} create national web corpora which are also largely composed of formal texts. They scrape websites which have national-top level domains (e.g. .au, .ca) corresponding to the countries. The authors conduct some frequency comparisons between each national corpus and are able to surface some terms unique to a country (e.g. Canadianisms like "heritage language" or "pot light"). This process is more automated but the resultant corpora similarly lack the informal variations we are studying.

Finally, the Twitter AAE corpus by \newcite{DBLP:journals/corr/BlodgettO17} most closely resembles the ones we are developing. It is a corpus of African American-aligned English tweets categorised based on probabilistic assumptions about tweet authorship. A tweet was placed in the African American-aligned corpus if it contained terms that were more frequently observed in tweets posted from areas with high concentrations of African-Americans. We make a comparable, albeit less statistically complex, geographic assumption in the creation of our corpus that tweets from a particular country are more likely to exhibit features of a given variety.

\subsection{Annotation Best Practice Guidelines}
The following list has been taken directly from \newcite{Blake2018InterannotatorAB} and reflects the guidelines we sought to follow in the creation of the annotation guidelines and throughout the annotation process:
\begin{enumerate}
    \item Annotate using tags at one level more finely than the research question requires.
    \item Provide clear rules and examples in which boundary cases are discussed in an annotation booklet.
    \item Develop, trial and require all annotators to complete a training course or session.
    \item Require annotators to reach a benchmarked standard.
    \item Mentor and provide constructive actionable feedback to annotators.
    \item Report inter-annotator agreement in sufficient detail to convince skeptical readers. 
\end{enumerate}

\section{Data} \label{data}
We use Twitter as a source for our data for two primary reasons. First, its international popularity and accessible API allows for simple data collection, due to the large volume of data which can be filtered based on location.\footnote{The Twitter API was only publicly-available until February 9, 2023.} Secondly, despite its written mode, the informal social media context means that some samples will be able to capture the differences between English varieties that would usually be found in casual spoken conversations. These differences include: code-switching, non-standard spellings to reflect phonetic variation and regional lexical markers. 

\subsection{Data Collection}
To minimize topical influence, our approach involves randomly selecting 100 tweets per day from those that were posted between January 1, 2022, and September 1, 2022. We explicitly consider 5 non-western cities: Accra, Islamabad, Manila, New Delhi, and Singapore. Our early observations show that tweets from the capital city capture linguistic variations better than tweets from smaller cities in the same country. We also collect tweets from New York and London as representatives of western English varieties. This results in a total of 170,800 tweets, with 24,400 tweets collected for each city. 

An attempt to filter tweets with less than 60\% English words shows that roughly 71.2\% tweets of our corpus satisfied the threshold. Upon our investigation, most filtered tweets are non-English. To ensure the quality of our corpus, we decide to retain only tweets that have at least 60\% English words. This will help us maintain a diverse range of code-switching and non-standard English terms in our corpus. This does not guarantee that non-English tweets are completely removed, as some non-English words may have the same spelling as English words.

\subsection{Linguistic Variation Analysis}
In our initial corpus evaluation, we identify orthographic variations, regional lexical markers, and syntactic differences in English varieties. Our analysis reveals differentiating features of these varieties, which in turn help us develop a comprehensive system of labeling for our annotation process.

\subsubsection{Orthographic Variation}

We find various instances of English phrases or words that are abbreviated, shortened, or altered. For example, "tbh" for "to be honest," "smh" for "shake my head," and "omg" for "oh my God" are frequently encountered abbreviations. Within Twitter's 280-character limit, such acronyms serve as linguistic shortcuts that enable users to convey their message efficiently. Similarly, English contractions like "gimme" (give me), "damit" (damn it), "needa" (need to), and "lemme" (let me) reflect the informal and conversational tone of tweets. Another important orthographic variation includes unconventional spellings and slang words like "bruh," "yey," "wassup," "lmao," and "hella." Some words are modified by adding repeated characters at the end, such as "hmmm," "omggg," and "plsss." These examples are used to express a specific tone, emphasis, or emotion, and have become increasingly prevalent among younger generations on social media. While such orthographic variations exist in both western and non-western tweets, tweets from New York and London tend to adhere more closely to formal language conventions. In fact, the percentage of formally-aligned English tweets in these cities is roughly 29\% higher than in other English-speaking communities we studied. 
\subsubsection{Lexical Variation}
To confirm the existence of lexical variation in our dataset, we calculate the percentage of non-English words present in the collected data. We first create a list of English words using several existing Natural Language Toolkit corpora, including wordnet2021, masc\_tagged, English stopwords, and Word Lists \cite{https://doi.org/10.48550/arxiv.cs/0205028}. For the tweets collected from all locations, we pre-process the data by removing digits, hashtags, mentions, links, punctuation, and non-alphanumeric characters (such as emojis). Our analysis shows that tweets from Accra, Islamabad, Manila, New Delhi, and Singapore had a higher percentage (67.9\%) of non-English words compared to tweets from London and New York (48.5\%). This is expected as tweets from London and New York are more likely to adhere to standard English. However, it is worth noting that some words were classified as non-English due to misspellings or named entities.

Some written varieties such as Singlish and Ghanaian English can be identified by the inclusion of additional non-English words. In Singlish tweets, discourse particle "lah" is often added in the end of the sentence. It has a range of pragmatic functions and serves a similar function to "\begin{CJK*}{UTF8}{gbsn}{了}\end{CJK*}" (le) in Mandarin -- an aspect marker, roughly translating to "already". The latter example is reflective of the significant Chinese influences on the language. In multiple tweets from Accra, the word "paa" is frequently used to add emphasis to a sentence, often appearing at the end. 
\subsubsection{Syntactic Variation}
Code-switching is commonly observed in multilingual locations like Accra, Islamabad, Manila, New Delhi, and Singapore. It involves the use, in English, of syntax structures and grammar rules from another language. This leads to  significant syntactic differences between (standard) American/British English and other English varieties. This phenomenon is particularly ubiquitous in informal social media settings, where speakers of multiple languages switch between them resulting in a high frequency of code-switched tweets. For example, Singapore has four official languages: English, Malay, Mandarin, and Tamil; the Philippines has two official languages: Filipino and English; Ghana's  official language is English, but local languages are widely spoken (including Twi, Fante, and Ewe). 

Another syntactic difference between English varieties and standard American and British English is reflected in verb conjugation. For example, consider the following tweet from Accra: 
\begin{displayquote}
\textit{"Bro! We be too relaxed for this side! We make am small then we relax"}
\end{displayquote}
The use of "be" instead of "are" or "are being" to indicate ongoing or habitual action is an example of non-standard English verb conjugation. Additionally, the use of "am" instead of "it" is another example of a non-standard usage.

\section{Annotation} \label{annotation}
The priority in designing annotation guidelines is ensuring that the classifications are (a) meaningful so annotators are able to confidently differentiate between labels whilst also coming to independent agreement about the classification of a text and (b) useful so the tweets in different categories contain extractable and informative features. The two primary researchers iteratively refine the annotation criteria, alternating between collaborative and independent labeling. They regularly apply inter-annotator agreement metrics to measure incremental improvements. The goal is to use class labels to isolate English varieties, by first distinguishing broadly between English and non-English words or sentences, and then distinguishing standard written English from English varieties. 
\subsection{Annotation Guidelines} \label{guidelines}
Below is a summarised version of annotation guidelines used by the primary researchers and the annotators. The labels do not correspond to their formal linguistic denotations. Rather, the labels succinctly conveyed the intended definition.
\subsubsection{Rules}
\begin{itemize}
    \item Treat existing standard named entities (person names, geopolitical entities, etc) as English when calculating the proportion of English
    \item Ignore tags, hashtags and emojis
    \item If a tweet meets the criteria for multiple labels in the list below, select the label that appears later in the list
\end{itemize}
\subsubsection{Labels}
\begin{enumerate}
    \item \textbf{Syntactic English} includes sentences which follow standard syntactic rules and are composed of standard English words.
    \item \textbf{Non-syntactic English} includes sentences of standard English words where at least one grammar rule is not followed (conjugation, word order, capitalisation, etc.)
    \item \textbf{Informal English} includes sentences with
    at least one word or term is misspelled (e.g. gnite) or is an informal or non-standard term, such as slang (e.g. yeet) or acronym (e.g. YOLO)
    \item \textbf{Code-switched} includes sentences with at least one region- or language-specific word such that someone from a different country and who speaks only English would not understand it. At least 40\% of the words are in English.
    \item \textbf{Incidental English} includes sentences where less than 40\% of the words are in English and there is at least one English word.
    \item \textbf{No English} includes sentences with exclusively non-English words.
\end{enumerate}
Some of the rules aim to indirectly isolate the various features of English varieties; for example, an analysis on the tweets composing rules 2 and 3, could highlight the morpho-syntactic and lexical differences of an English variety while rule 4 targets tweets which incorporate use of other languages.

Annotators are provided a simple interface consisting of the raw original tweet and a cleaned version consisting of the uncapitalised tweet with hashtags, user tags, hyperlinks and punctuation removed. The raw tweet is necessary when assessing the syntactic correctness of the tweet while the cleaned tweet allows for a simpler way to identify and count any non-English words.

\subsection{Inter-annotator agreement benchmarks}\label{benchmarks}
After the final iteration of the annotation guidelines, the primary researchers independently annotated 500 novel tweets from Singapore. 

\begin{figure}[!h]
\begin{center}
\includegraphics[scale=0.55]{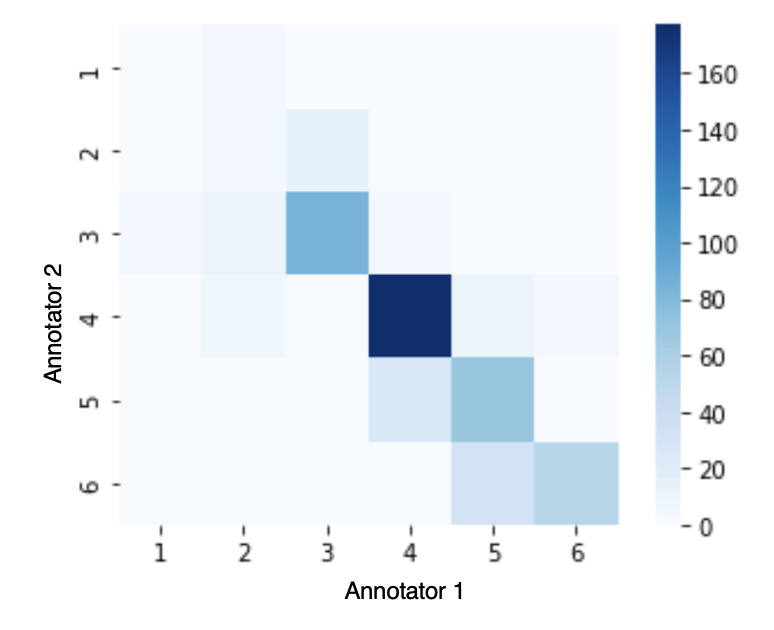} 
\caption{Confusion matrix -- 500 
Singapore tweets}
\label{heat_map}
\end{center}
\end{figure}

Using metrics for independent, nominal categories, the percent agreement between the researchers is 77\% and the Cohen’s Kappa coefficient - a commonly used statistical measure of inter-rater reliability that accounts for the possibility of agreement by chance \cite{GISEV2013330}  - is 0.69 which is considered `substantial' \cite{10.2307/2529310}. Given the large number of categories and their relative complexity, these scores can be seen as strong indicators that the definitions allow for meaningful differentiation. 

One important observation in the confusion matrix (Figure \ref{heat_map}) is that the disagreement is concentrated around the diagonal. This suggests that adjacent categories resemble each other much more closely than distant ones. This makes sense for categories 4-6 because there are quantitative boundaries (i.e. proportion of English). The relationship between categories 1-4 is less obvious except for the fact that later categories are chosen if criteria is met for multiple categories, meaning that there are stricter requirements for lower-ranked categories. As soon as an annotator detects the presence of one category’s defining characteristics, they could not classify the tweet into an earlier category. This trend is confirmed by the decreasing number of tweets classified from \textit{Code-switched} to \textit{Syntactic English}. The relationship between adjacent categories points to an implicit ordinality, which reflects our intention to create labels that indirectly measures distance from formal standard English. 

For ordinal categories, inter-annotator agreement can be additionally and appropriately measured by correlation coefficients that weigh the similar categories as an indication of partial agreement \cite{GISEV2013330}. Applying the quadratically-weighted Cohen’s Kappa (0.87) and the Kendall coefficient of concordance (0.86), we observe very strong agreement between annotators. As a benchmark for annotators, we required a quadratically-weighted Cohen’s Kappa exceeding 0.8 before allowing them to independently annotate.

\subsection{Annotation training}\label{training}
We recruited 5 students who are nationals of the countries we aimed to study. A brief summary of their profiles can be seen in Table \ref{train}. All students are completing their tertiary education in English and are thus fluent in English. They are also speakers of dominant local languages from their respective countries. While it was not strictly necessary for annotators to understand the content of each tweet, this annotation process was an opportunity to gather informed insights about tweet topics and linguistic nuances that individuals of diverging backgrounds could identify. 

Each annotator was guided through a training session in which they were briefed about the research goals and provided with a comprehensive explanation of the annotation guidelines. They were guided through a tutorial gave 25 examples of Singaporean tweets for each category and instructions for handling edge cases. We then provided them with 100 pre-labelled tweets from Singapore that served as a qualification task for which they needed to achieve a quadratically-weighted Cohen’s Kappa exceeding 0.8. While we had prepared a further 200 pre-labelled tweets for an additional two rounds of iterative feedback before disqualification, all 5 annotators met our benchmark on their first attempt (Appendix \ref{appendixA}), which is further evidence of the distinctiveness of each category. 

We asked the annotators to list any ambiguities they faced in this process and we provided individualised feedback to further improve the quality of their annotations before allowing them to move on to annotating their assigned tweets. Subsequently, we sampled the collected and filtered tweets into batches of 100 and distributed the sets corresponding to their country. The number of batches completed varied for each annotator depending on their availability. By their last batches, annotators were able to label around 200 tweets per hour on average. In future, each tweet should be classified by independent annotators to further ensure reliability. 

\subsection{Annotator observations}
We asked each participant to do a short voice note recording of any observations they made through the process of annotation. Many highlighted demographic and topical trends. For example, all indicated that local politics was a common subject; annotator 4 frequently noticed expressions of patriotism or praise for the Indian Prime Minister. They also repeatedly saw tweets about religion and cricket which are in line with the common national interests. Annotator 1 noticed that the political tweets in the Philippines tended to be critical of the ruling party and that users more frequently discussed daily activities with their followers. These reports verify that Twitter serves differing social functions in different countries \cite{poblete2011all}, which means that subject matter could be used as an indicator for tweet origin.

The annotators were also asked about their experience with the annotation guidelines. Some expressed confusion about what constitutes a syntactic English sentence when deciding whether a tweet constitutes \textit{Syntactic} or \textit{Non-syntactic} English. It was perhaps unfair to expect annotators to know formal morpho-syntactic rules, especially when most language users can rarely articulate the reason a sentence is well-formed. Given that there were proportionally few examples of tweets labeled \textit{Syntactic English}, it seems appropriate to merge these categories in future. 

In contrast, many annotators indicated that the \textit{Code-switched} category lacked granularity and felt like they were `lumping' linguistically dissimilar tweets into the same bucket. For example, annotator 5 said they used the \textit{Code-switched} label for tweets in which users switched between English and a local language and for tweets which used Ghanaian Pidgin English (Broken English). Similarly, annotator 1 distinguished between (i) tweets where only one language (English or Tagalog) was used in each of multiple composing sentences and (ii) tweets where Tagalog words were interwoven in predominantly English sentences. Both types were labeled \textit{Code-switched} according to the criteria. This suggests that the \textit{Code-switched} category could be broken up into even more descriptive categories, especially given that it constituted the plurality of labeled tweets. However, since each annotator identified different distinguishing linguistic features - e.g. lexicon for Ghanaian tweets and sentence composition for Filipino tweets - appropriate new categories may have to be specific and contextually-appropriate for each country. 

\subsection{Corpus Statistics}\label{stats}
\begin{figure}[!h]
\begin{center}
\includegraphics[scale=0.5]{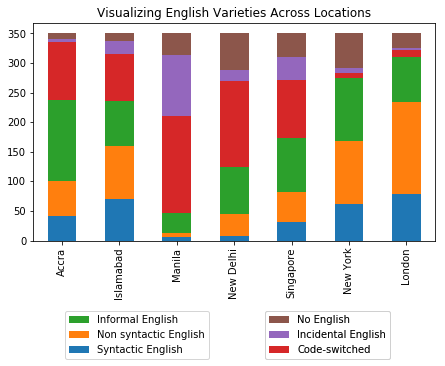} 

\caption{Overall label distribution of 350 tweets in each of the locations: Accra, Islamabad, Manila, New Delhi, Singapore, New York and London}
\label{overall}
\end{center}
\end{figure}
Our manually labeled subset of 3564 tweets constitutes approximately 7.2\% of the total corpus. From these, we  selected 350 labeled tweets with at least 70\% English from each location to examine the category distribution. Figure \ref{overall} shows that \textit{Code-switched} and \textit{Informal English} are the dominant categories in tweets from Accra, Islamabad, and Singapore, while \textit{Code-switched} is clearly the most significant category for Manila and New Delhi tweets. The variations in category distributions across these locations are evidence for linguistic diversity among English varieties in our corpus. Specifically, the presence of code-switching in tweets from these locations highlights the blending of English with local languages, resulting in a more diverse and unique dataset that cannot be captured in standard formal English texts. Furthermore, the dominance of \textit{Code-switched} tweets in Manila and New Delhi further emphasizes the prevalence and importance of multilingualism in these regions. 

While it is expected that tweets from New York and London are more likely to resemble standard English, we still observe a higher percentage of \textit{Non-syntactic English} and \textit{Informal English} tweets. This may be attributed to the informal nature of social media platforms such as Twitter, where users tend to communicate in a more casual and less structured manner. The \textit{Code-switched} category was rarest in these two locations. This is not surprising since code-switching typically occurs in multilingual contexts, and thus the predominance of monolingual English speakers in New York and London may account for the lower incidence of code-switching.

\section{NLP Tools on English Varieties}\label{langid}
We evaluate three language identifiers - \textit{langid.py}, \textit{spaCy-langdetect}, and \textit{Google Translate API} - for detecting English vs. non-English tweets in our corpus. In our annotation framework, tweets in categories 1-3 should be classified as English without question. For category 4, code-switching, we randomly sampled 100 tweets from Accra, Islamabad, Manila, New Delhi, and Singapore, and manually determined the proportion of English words, including both standard and informal language. Our analysis reveals that 82\% of the \textit{Code-switched} tweets had at least half of their content in English. Thus, we consider \textit{Code-switched} tweets as English.

When evaluating these tools on 350 tweets from each city, we find that their accuracies are much higher for American and British English than for other English varieties. Our results are tallied up in Table \ref{tab1} and \ref{tab2} (Appendix \ref{appendixB}). The differences are 21.82\%, 19.77\%, and 32.58\% for
\textit{langid.py}, \textit{spaCy-langdetect}, and \textit{Google Translate}. Even when we use a more conservative approach and do not consider tweets in \textit{Code-switched} category as English, the tools demonstrate higher accuracy for western English than non-western English. The differences in accuracy scores between western and non-western English are 12.43\%, 7.33\%, and 8.93\% respectively. These gaps in accuracy between English varieties highlight ethical implications. Such NLP tools are built and trained on large datasets of American and British English, which leads to bias and inaccuracies when applied to other English varieties. This can result in discrimination against individuals who speak non-standard varieties of English, e.g., in automated hiring processes.

\section{Future Work} \label{future_work}
While the initial annotation process was labor-intensive, it was essential for us to develop a thorough annotation framework. This serves as a potentially valuable resource in the continuing labeling process, ensuring that the labeled data is consistent and of high quality. Nevertheless, the limited amount of labeled data in our corpus poses a significant challenge to achieve sufficient coverage and accuracy in many NLP tasks. To address this issue, we plan to explore active learning techniques or large language models (LLMs) such as GPT-3 to expand our corpus. The labeled portion of our corpus will serve either as seeds to train a machine learning model that can then identify similar, yet unlabeled, data points in the remaining corpus, or as a baseline to evaluate annotations produced by LLMs. Moreover, our corpus experiments focus on a fundamental NLP task: language identification, specifically identifying English vs. non-English tweets. We demonstrate the biases in pre-trained language models towards non-western English varieties. However, we recognize the need to broaden the scope of our investigation to other NLP tasks, with the goal of developing more robust, inclusive and accurate NLP models. For example, our corpus could be useful in training sentiment analysis models to analyze the sentiments of writers from different linguistic backgrounds. Finally, we are committed to expand our corpus to include a wider range of global English-speaking communities. While our current corpus covers several countries in Asia and Africa, we recognize that there are many more English varieties around the world with linguistic characteristics different from those in our corpus. We believe that this continued expansion of our corpus will benefit researchers studying subfields of linguistics like sociolinguistics and corpus linguistics and enable the development of robust NLP models that are better suited to the needs of diverse linguistic communities.

\section{Conclusion}
We have presented a diverse tweet corpus of English varieties and an annotation framework to label tweets. We analyse linguistic indicators of these English varieties and demonstrate that, despite a superficially independent relationship between some `adjacent' labels, the classifications exist along a spectrum which intuitively measures distance from formal, standard English. We train several linguistically-diverse and geographically-appropriate individuals to annotate a collective 3564 tweets. Our experiments also show that there exists bias towards English varieties in off-the-shelf language identification tools when evaluated on our diverse corpus. 

\section*{Limitations}
We acknowledge some limitations of Twitter, notably that its usage is concentrated among wealthy, white and western individuals and even within non-western countries, the majority of active users is typically younger and more educated \cite{blank2017representativeness}. This means that our corpus may be skewed towards more privileged populations of certain demographics and thus not accurately represent the linguistic diversity of English-speaking communities, especially those that are underserved. Nevertheless, we believe that our corpus is still a valuable contribution towards inclusivity in NLP as we increase the representation of English varieties beyond American and British English. Moreover, we point out the issue of existing bias in off-the-shelf language identification models, but we do not directly address it. This is because such a task would require a much more substantial amount of annotated training data than we currently have. As outlined in Section \ref{future_work}, we plan to scale our labeled data in a more efficient and automated manner, which will enable us to better address this limitation.

\bibliographystyle{acl_natbib}
\bibliography{acl_latex}

\onecolumn
\appendix
\section{Annotation Qualification Task} \label{appendixA}
\begin{table*}[h!]
\centering
\begin{tabular}{ |c|c|c| }
\hline \textbf{Country of Origin (Languages Spoken)}& \textbf{Datasets labelled} & \textbf{Weighted Cohen’s Kappa}\\ 
\hline\hline The Philippines (Tagalog and dialect) & Philippines & 0.86\\
\hline Pakistan (Urdu) & Pakistan & 0.87\\
\hline Pakistan (Urdu) & Pakistan and India & 0.82\\
\hline India (Hindi and Urdu) & India & 0.85\\
\hline Ghana (English, Twi, Hausa, Pidgin) & Ghana & 0.84\\
\hline
\end{tabular}
\caption{Results of annotation qualification task in annotation training}
\label{train}
\end{table*}

\section{Language Identification on Tweets of English Varieties} \label{appendixB}
\begin{table*}[h!]
  \scalebox{0.95}{\centering
  \begin{tabular}{|c|c|c|c|c|c|c|c|}
   \hline \textbf{Language Identifier} & \textbf{Accra} & \textbf{Islamabad} & \textbf{Manila} & \textbf{New Delhi} & \textbf{Singapore} & \textbf{New York} & \textbf{London}\\ 
    \hline
    \hline langid.py & 72.32 & 69.62 &  60.19 & 62.45 & 71.32 & 87.32 & 90.68\\
    \hline spaCy & 77.38 & 85.76 & 55.45 &  73.98& 77.57  & 92.25 & 95.34\\
    \hline Google Translate API &80.95 &81.86 & 33.65& 58.74& 61.03 &95.42& 96.27\\
    \hline
  \end{tabular}}
  \caption{Percentage of the 350 tweets in each English variety classified as English}
  \label{tab1}
\end{table*}

\begin{table*}[h!]
  \centering
  \begin{tabular}{|c|c|c|c|}
   \hline \textbf{Language Identifier} & \textbf{American \& British English} & \textbf{Other English varieties} & \textbf{Difference}\\ 
    \hline
    \hline langid.py & 89.00 & 67.18 & \textbf{21.82}\\
    \hline spaCy  & 93.80& 74.03 & \textbf{19.77}\\
    \hline Google Translate API  & 95.85 & 63.27 & \textbf{32.58}\\
    \hline
  \end{tabular}
  \caption{Average accuracy of western English and non-western English varieties}
  \label{tab2}
\end{table*}

\end{document}